\pdfoutput=1

\documentclass[11pt]{article}

\usepackage[]{acl}

\usepackage{times}
\usepackage{latexsym}
\usepackage{graphicx}
 \usepackage{booktabs}
 \usepackage{hyperref}
 \usepackage{caption}
\usepackage{subcaption}

\usepackage[T1]{fontenc}

\usepackage[utf8]{inputenc}

\usepackage{microtype}

\title{UKP-SQuARE: An Interactive Tool for Teaching Question Answering}

\author{Haishuo Fang, Haritz Puerto, Iryna Gurevych \\
Ubiquitous Knowledge Processing Lab (UKP Lab), \\Department of Computer Science and Hessian Center for AI (hessian.AI), \\Technical University of Darmstadt \\
\url{www.ukp.tu-darmstadt.de}
}

\begin{document}
\maketitle
\begin{abstract}
    The exponential growth of question answering (QA) has made it an indispensable topic in any Natural Language Processing (NLP) course. Additionally, the breadth of QA derived from this exponential growth makes it an ideal scenario for teaching related NLP topics such as information retrieval, explainability, and adversarial attacks among others. In this paper, we introduce UKP-SQuARE as a platform for QA education. This platform provides an interactive environment where students can run, compare, and analyze various QA models from different perspectives, such as general behavior, explainability, and robustness. Therefore, students can get a first-hand experience in different QA techniques during the class. Thanks to this, we propose a learner-centered approach for QA education in which students proactively learn theoretical concepts and acquire problem-solving skills through interactive exploration, experimentation, and practical assignments, rather than solely relying on traditional lectures. To evaluate the effectiveness of UKP-SQuARE in teaching scenarios, we adopted it in a postgraduate NLP course and surveyed the students after the course. Their positive feedback shows the platform's effectiveness in their course and invites a wider adoption.
\end{abstract}
\section{Introduction}
Question Answering (QA) is one of the overarching research topics in Natural Language Processing (NLP). QA pipelines have been developed to address different types of questions, knowledge sources, and answer formats, including extractive, abstractive, knowledge base, multiple-choice, generative, and open-domain QA. Such a massive number of QA systems and relevant NLP techniques are making QA lectures more important in NLP courses. However, despite QA being an application-oriented topic (e.g., chatbots, virtual assistants, etc.), classes are usually theoretically driven. Thus, in this paper, we propose the use of the UKP-SQuARE platform as a tool for QA education. This platform integrates most QA formats, popular models, datasets, and analysis tools, such as explainability, adversarial attacks, and graph visualizations.

Compared with conventional teacher-led classes, we propose a learner-centered class following the flipped classroom \citep{bishop2013flipped} with UKP-SQuARE as the driving tool of the lecture. This tool provides an interface for users to interact with different QA models and analysis tools. Therefore, students can actively learn about QA systems and get hands-on experience by interacting with models on the platform. Concretely, students can flexibly compare multiple architectures that model different QA formats, analyze their outputs with explainability tools, and even analyze their robustness against adversarial attacks. Prior studies have shown that flipped classroom lectures improve the learning process of students in programming courses \citep{interactive_learning}. Thus, we believe that teaching and learning QA through a live demo with this platform can also make NLP lectures more engaging, drawing students' attention, and interest in the topics.

To investigate the effectiveness of UKP-SQuARE in QA education, we adopted it for the first time in a postgraduate NLP course\footnote{Master's level course} and conducted a survey afterward. The positive feedback from the students encourages us to continue adopting this platform and education method in more NLP courses. The contributions of this paper are: i) a novel interactive learner-centered methodology to teach QA and relevant NLP topics, ii) extending the UKP-SQuARE platform for teaching QA, and iii) the design of a syllabus for interactive QA lectures.

\section{UKP-SQuARE}

UKP-SQuARE~\citep{baumgartner-etal-2022-ukp,sachdeva-etal-2022-ukp,puerto2023ukp} is an extendable and interactive QA platform that integrates numerous popular QA models such as deeepset's roberta-base-squad2\footnote{\url{https://huggingface.co/deepset/roberta-base-squad2}}, SpanBERT \citep{joshi-etal-2020-spanbert} for HotpotQA, and QAGNN \citep{yasunaga-etal-2021-qa}. It provides an ecosystem for QA research, including comparing different models, explaining model outputs, adversarial attacks, graph visualizations, behavioral tests, and multi-agent models.
In addition, this platform provides a user-friendly interface\footnote{\url{https://square.ukp-lab.de/}} that enables users to interact. Users can run available models, deploy new ones, compare their behaviors, and explain outputs.

\section{Learning Question Answering with UKP-SQuARE}

In this section, we present the syllabus of a lecture focused on QA and relevant NLP topics that use the platform UKP-SQuARE following the flipped classroom methodology ~\cite {bishop2013flipped}. The flipped classroom is an effective learner-centered educational methodology in which students study pre-recorded lectures and materials in advance to engage in more interactive and collaborative learning activities in class. UKP-SQuARE can be the driving tool for the flipped classroom in QA education. With our platform, lecturers can introduce the topics by interacting with the students and then proceed to an in-depth explanation of the technical details behind the methods of each topic. We propose dividing the lecture into three topics in the QA field: basic QA concepts, trustworthy QA, and multi-agent QA systems. With these topics, students can learn about QA and related NLP topics such as information extraction, explainability, adversarial attacks, and multi-agent systems.

\subsection{Learning Basic QA Components}
QA systems include two main components, i.e., Readers and Retrievers. Readers are QA models responsible for obtaining answers from the context retrieved by retrievers. In UKP-SQuARE, students can easily learn various readers (QA models) within different QA formats and information retrieval techniques via interacting with the interface.
\subsubsection{Contrasting Different QA Formats}
\begin{figure}[!htb]
    \centering
    \begin{subfigure}{\columnwidth}
        \includegraphics[width=\textwidth]{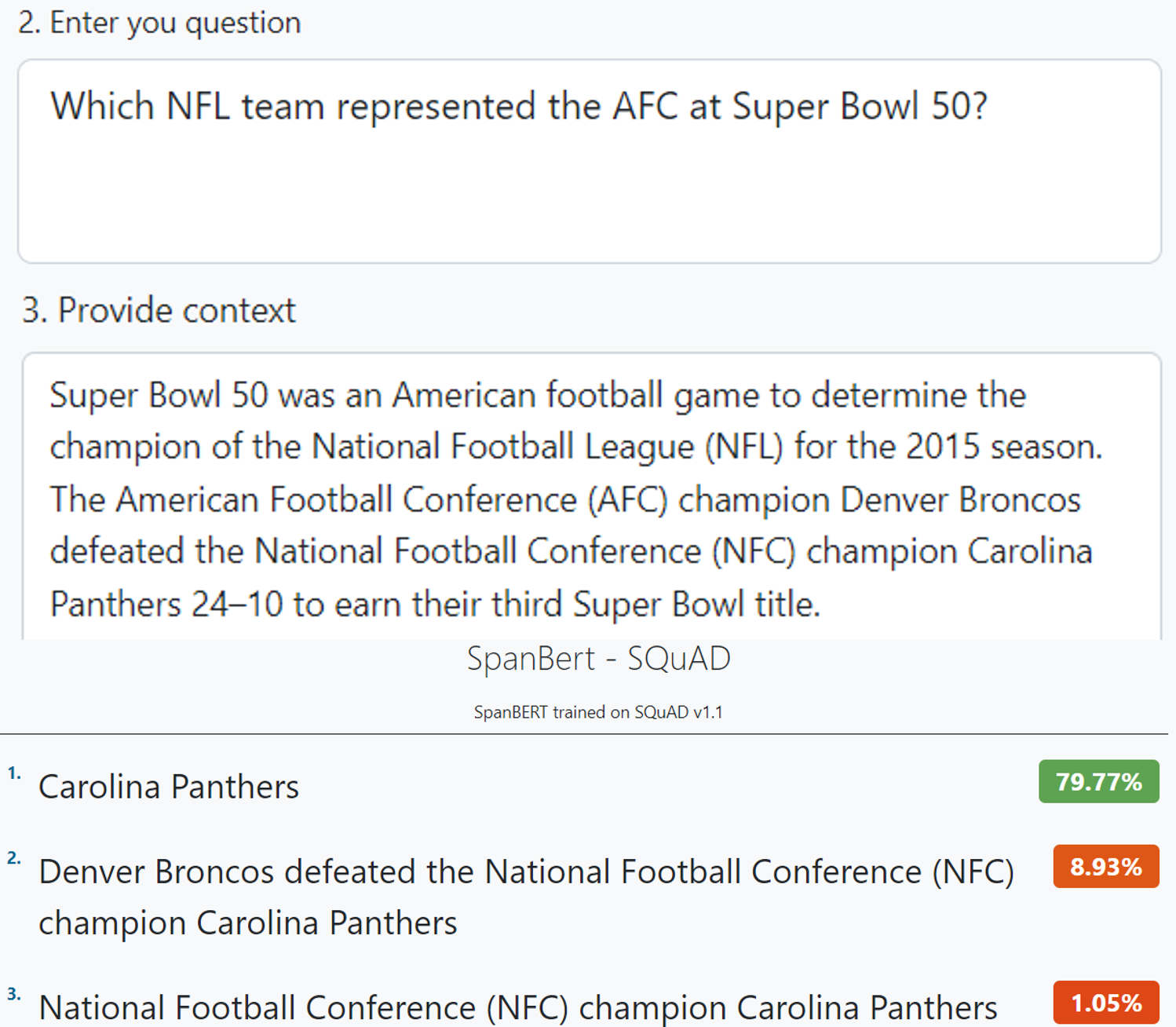}
        \subcaption{A span-extraction QA model}
    \end{subfigure}
    \begin{subfigure}{\columnwidth}
        \includegraphics[width=\textwidth]{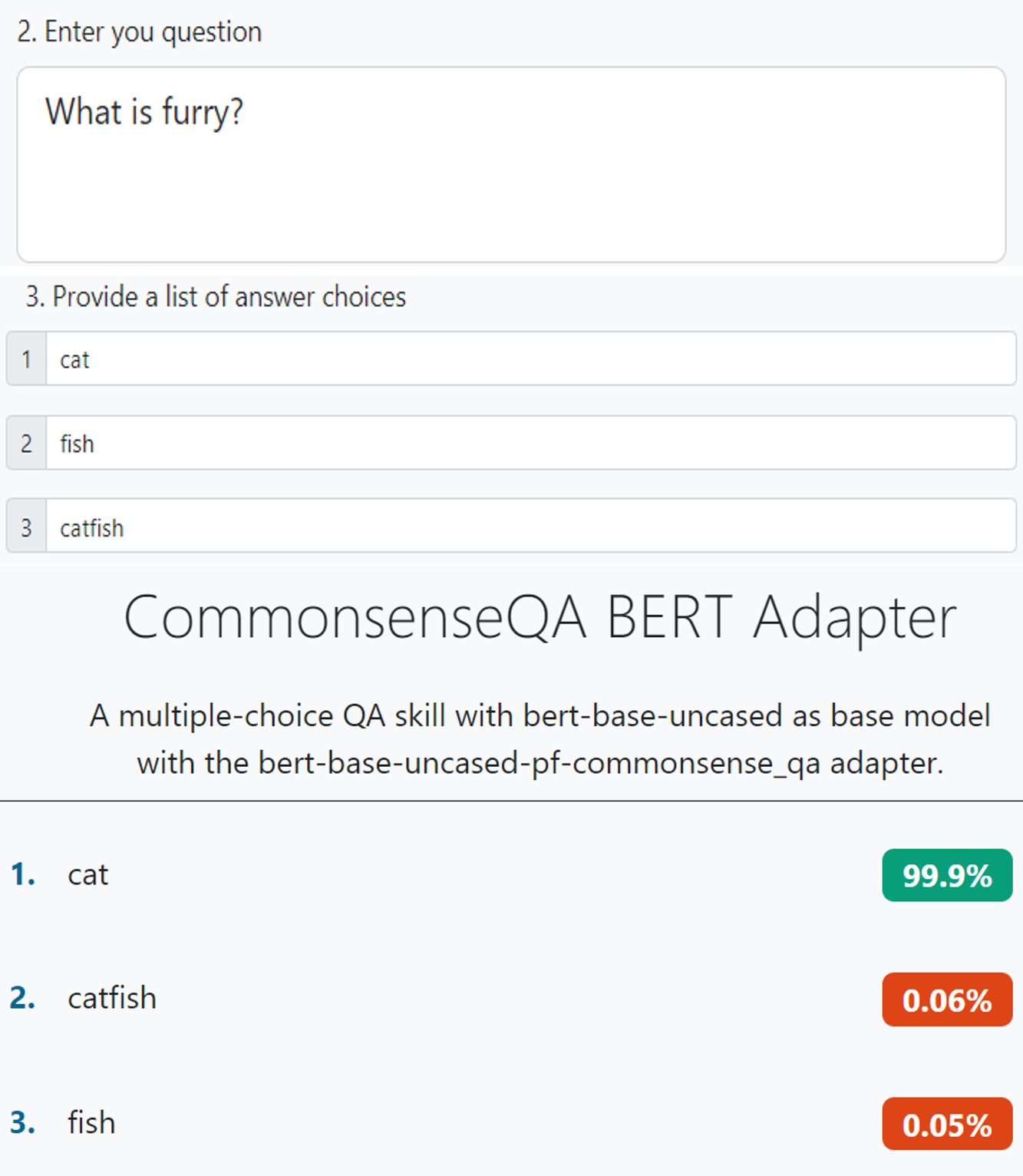}
        \subcaption{A multiple-choice QA model}
    \end{subfigure}
    \caption{Different QA formats in UKP-SQuARE}
    \label{fig:constrast_models}
\end{figure}

With UKP-SQuARE, students can get first-hand experience by interacting with multiple models on our platform. The home readings would include descriptions of the main QA datasets and their baselines. In class, the lecturer can show the different QA formats with real demonstrations of the models and explain on the fly the architectural differences needed to model each QA format. An example is shown in Figure~\ref{fig:constrast_models} where a span-extraction QA model, i.e., Span-BERT, and a multiple-choice QA model, i.e., CommonsenseQA model are presented to show the difference between these two QA formats. Such interactions can make theoretical explanations of the architectures easier to digest and, therefore, the class more engaging.

\subsubsection{Learning Information Retrieval}
\begin{figure}
    \includegraphics[width=\columnwidth]{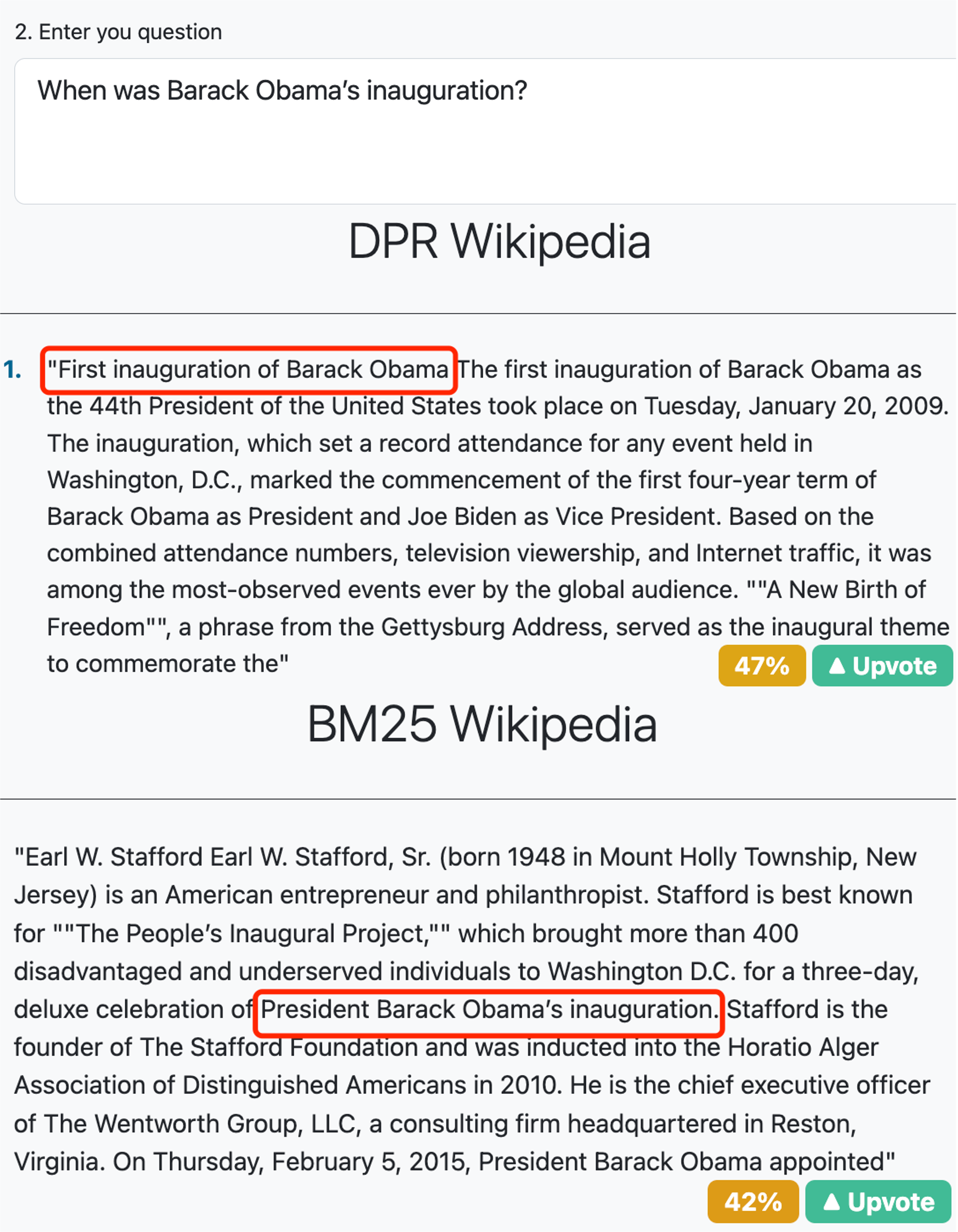}
    \caption{Example of difference between using BM25 retriever and DPR retriever. The red boxes represent keywords in the retrieved passages}
    \label{fig:ir}
\end{figure}

To learn Information Retrieval (IR) methods, the user interface of UKP-SQuARE offers a compelling approach to help students differentiate between different IR methods, e.g., lexical retrieval and semantic retrieval, and understand how they affect the final performance of QA models. The home readings would include book chapters or slides describing the main IR methods such as TF-IDF \citep{tfidf}, BM25 \citep{robertson1995okapi}, Sentence-BERT \citep{reimers-gurevych-2019-sentence}, and Dense Passage Retrieval \citep[DPR;][]{karpukhin-etal-2020-dense}. Like the above section, the lecturer can guide students to find the difference between lexical retrieval (e.g., BM25) and semantic retrieval (e.g., DPR) via playing with UKP-SQuARE by themselves. As shown in Figure~\ref{fig:ir}, for the question \textit{When was Barack Obama’s inauguration?}, the BM25 retriever returns a passage covering all keywords but irrelevant to the question, while the DPR retriever returns the correct document, which contains the answer to the question. By providing this example in class, students can easily understand that DPR retrieves semantically similar passages while BM25 only retrieves passages that contain the query tokens and, thus, may retrieve unrelated passages. This could be further explored by comparing two open-domain QA models implementing these retrieval methods and the same reader model to demonstrate the error propagation due to irrelevant passages. This active learning method can prevent the issue of students losing attention that commonly occurs in traditional lectures \cite{felder2003designing}. 

\subsection{Learning Trustworthy QA Systems}
In addition to learning basic QA components, it is important to understand how to identify and evaluate trustworthy QA systems. This involves several related NLP topics, such as explainability, transparency, and robustness. UKP-SQuARE provides such analysis tools to facilitate students' learning process of trustworthy QA systems.
\subsubsection{Explainability Methods}
The exponential adoption of AI is pushing regulators to adopt policies to regulate its use. One of the key points they aim to address is the explainability of these methods to make AI safer\footnote{\url{https://digital-strategy.ec.europa.eu/en/policies/european-approach-artificial-intelligence}}. Thus, it is of utmost importance to include explainability methods on AI courses in Universities. In terms of the explainability of QA models, UKP-SQuARE includes BertViz \cite{vig2019bertviz} and a suite of saliency map methods to facilitate the understanding of the model's decision-making process. Saliency maps employ attribution-weighting techniques such as gradient-based \cite{DBLP:journals/corr/SimonyanVZ13,DBLP:conf/icml/SundararajanTY17} and attention-based \cite{jain-etal-2020-learning,serrano-smith-2019-attention} methods to determine the relative importance of each token for the model prediction. The descriptions of these methods would form part of the home readings and to make the classes more active, the class would be driven by real examples of saliency maps using our platform and their interpretation. In this way, students can learn how to explain the output of a QA model based on saliency maps.

\begin{figure}
    \centering
    \includegraphics[width=\columnwidth]{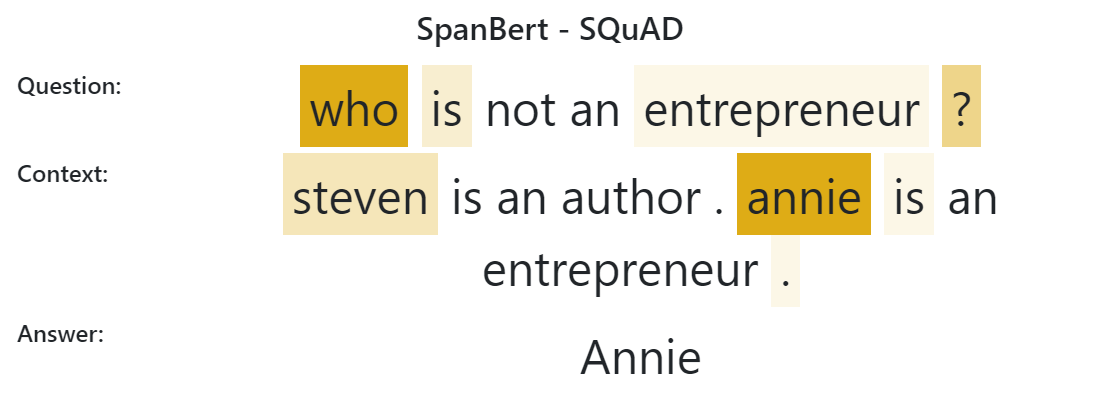}
\caption{An attention-based saliency map of a question in UKP-SQuARE.}
    \label{fig:attention}
\end{figure}

An example of a saliency map is shown in Figure~\ref{fig:attention}. The color level of the highlighted text reflects its importance for the answer. As we can see, \textit{of what celestial body?} is the most important part of the question, while \textit{sun} gets the most attention in the context, which is the final answer. This means the model successfully understands the main point of the question and can link them to the context. Making this type of interpretation can help students identify potential problems or biases in the models.

\subsubsection{Behavioral Tests in QA models}
The next important component in trustworthy QA is behavioral tests of models. Machine learning models do not throw errors as regular software programs. Instead, an error in machine learning is usually an unwanted behavior, such as a misclassification that may pass inadvertently to a person \citep{ribeiro-etal-2020-beyond}. This makes testing machine learning models challenging. To simplify the behavioral analysis of machine learning models, \citet{ribeiro-etal-2020-beyond} proposes \textit{CheckList}, a list of inputs and expected outputs that aims to analyze general linguistic capabilities and NLP models mimicking the unit tests in software engineering. The integration of \textit{CheckList} into UKP-SQuARE offers a simple method to analyze the performance of QA models beyond traditional benchmarks, such as MRQA tasks \citep{fisch-etal-2019-mrqa}.

\begin{figure}
    \centering
    \includegraphics[width=\columnwidth]{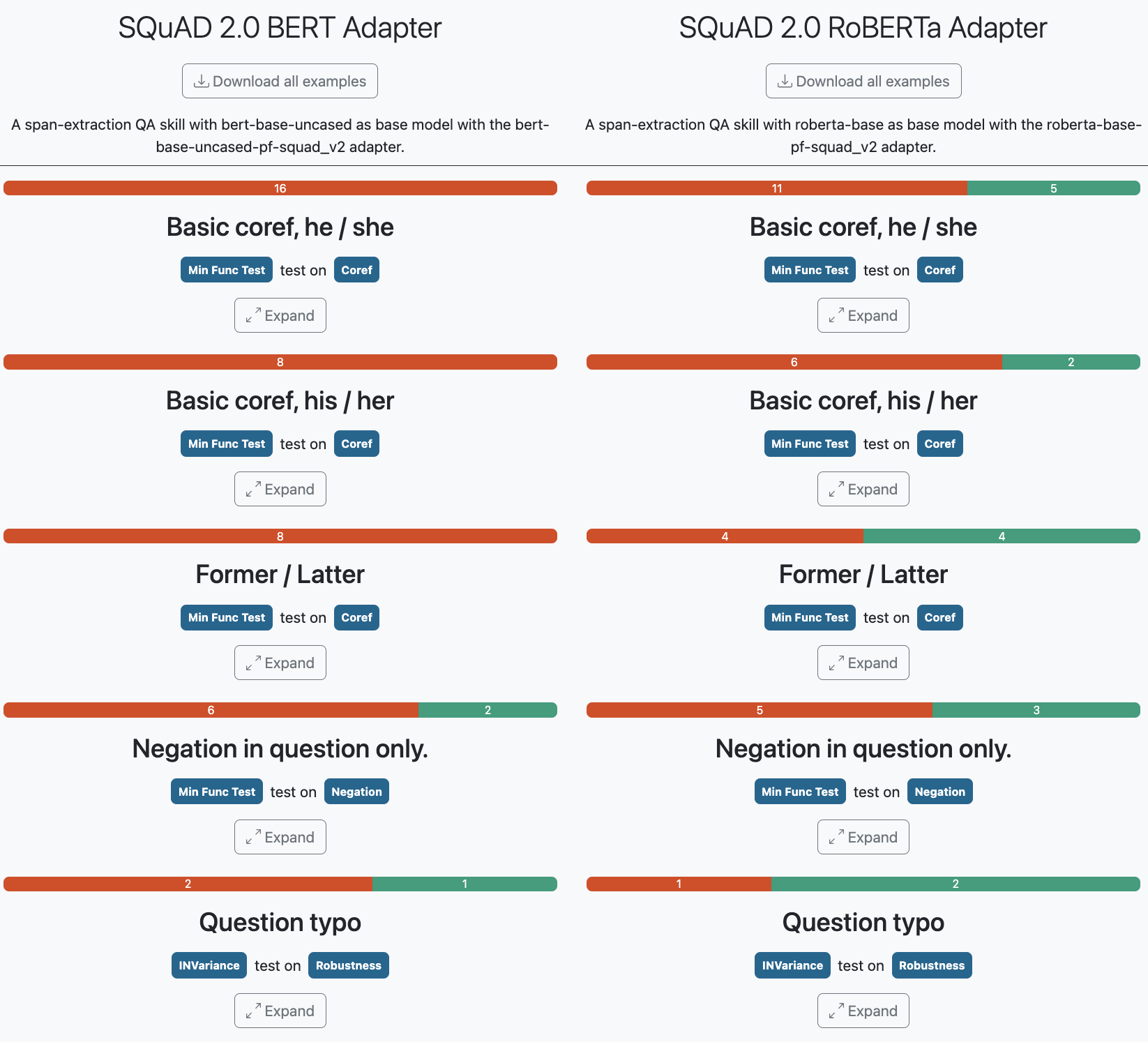}
    \caption{The result of running CheckList for SQuAD 2.0 RoBERTa Adapter and BERT Adapter. The number of failed and succeeded test cases are highlighted in green and red.}
    \label{fig:checklist}
\end{figure}

As illustrated in Figure~\ref{fig:checklist}, we test the SQuAD 2.0 RoBERTa Adapter and SQuAD 2.0 BERT Adapter using the CheckList in which multiple NLP capabilities are tested like coreference, negation, and robustness. As we can see SQuAD 2.0 BERT Adapter performs worse than RoBERTa Adapter in the above dimensions.
Such an example can be used by the lecturer in class to introduce the idea of behavioral tests on the fly.
In addition, the behavioral tests of UKP-SQuARE can be used to foster the students' analytical skills. A potential assignment could be to train a QA model and deploy it on our platform to analyze it with the provided ecosystem of QA tools. In particular, thanks to the behavioral tests in UKP-SQuARE, students can provide a deeper analysis of their model based on the quantitative results of their test set and a qualitative analysis based on the behavioral test results.

\subsubsection{Adversarial Attacks}
Policymakers are also designing a regulatory framework that guarantees users that their AI models are resilient to adversarial attacks\footnote{See footnote 3}. Therefore, AI curriculums should also include adversarial attacks to prepare students for these new regulations.

UKP-SQuARE provides tools to conduct adversarial attacks, such as HotFlip \cite{ebrahimi-etal-2018-hotflip}, input reduction \cite{feng-etal-2018-pathologies}, and sub-span \cite{jain-etal-2020-learning}. Thus, the home readings should include a theoretical introduction to these methods. Then, the lecture would use the platform to exploit the interactive nature of adversarial attacks. In particular, the need to analyze examples to understand different types of attacks makes this part of the topic especially practical. Therefore, the lecturer can introduce the topic through UKP-SQuARE and delve deeper into the technical details afterward.

\begin{figure}
    \centering
    \includegraphics[width=\columnwidth]{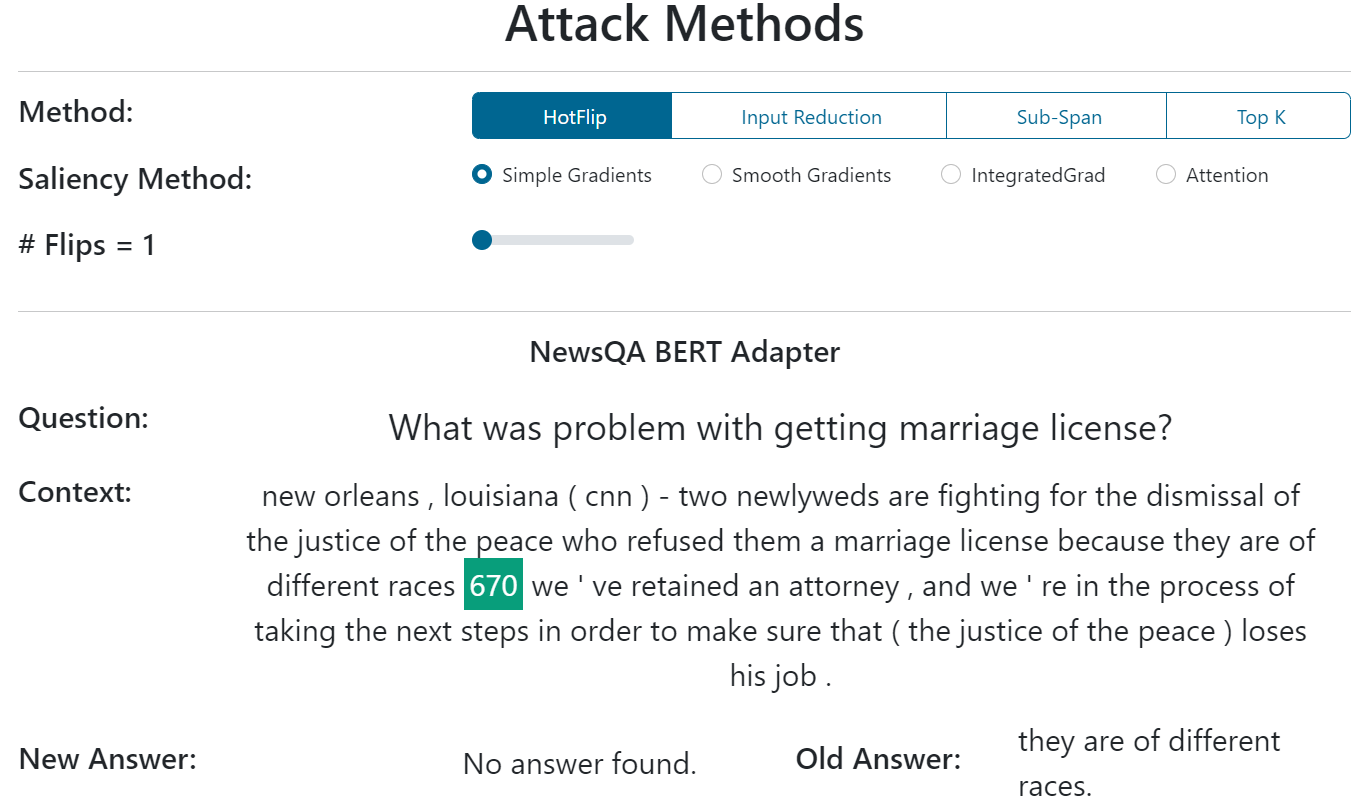}
    \caption{A HotFlip example where only flipping \textit{. (full stop)} to \textit{670} changes the answer.}
    \label{fig:hotflip}
\end{figure}
An exemplary case is that students can attack real models with examples by tuning different parameters, such as the number of flips in HotFlip, to see how the output changes when they subtly change the input data. In Figure~\ref{fig:hotflip}, only flipping \textit{. (full stop)} to \textit{wore} can directly change the answer. In class, a small experiment can be set up by lecturers in which students need to manually manipulate the input to see if it can trick the model into making incorrect answers and compare it with adversarial attack tools to deepen their understanding of those adversarial attacks and the importance of building up trustworthy QA systems.

\subsubsection{Graph-based QA Models}

Knowledge Graph Question Answering (KGQA) systems can have strong explanatory power thanks to the reasoning paths that can be extracted from the graph. Such transparency can enhance the interpretability and trustworthiness of the system. UKP-SQuARE currently offers QA-GNN ~\cite{yasunaga-etal-2021-qa}, a KGQA model that makes use of ConceptNet~\citep{speer2017conceptnet}, and provides a visualization interface to explore the subgraph used by the model. 

Although a reasoning path in a graph may provide a clear explanation of a model's prediction, we believe that interpreting graph-based models is not straightforward because, usually, that path contains many irrelevant nodes and edges that may obscure the actual reasoning of the model. Thus, we propose to teach KGQA models with real examples of graphs. In this way, the lecturer, or even the students themselves, have to show the process of cleaning the graph to obtain and interpret the reasoning path. This process would be much more valuable for the future endeavor of the students than using a set of slides with examples of preprocessed clean graphs because they will be able to reproduce what they learn in real-use cases in companies.

\begin{figure}
    \centering
    \includegraphics[width=\columnwidth]{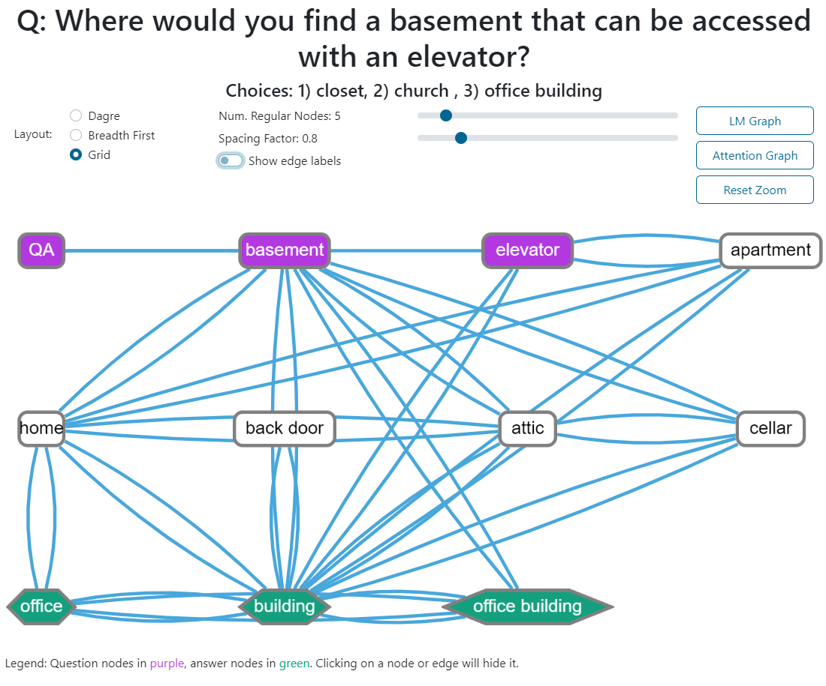}
    \caption{A visualized reasoning graph of the question \textit{Where would you find a basement that can be accessed with an elevator?}}
    \label{fig:qagnn}
\end{figure}

\subsection{Learning Multi-Agent Systems}
Lastly, the current progress in QA is pushing toward creating robust models across multiple domains. To do this, there are two types of approaches: multi-dataset models and multi-agent models. While the former aims to train a single architecture on multiple datasets, the latter does the opposite. It trains multiple models (agents) on single datasets and combines the agents. UKP-SQuARE is compatible with both approaches; therefore, it is an ideal platform to teach them. 

\begin{figure}[!htb]
    \includegraphics[width=\columnwidth]{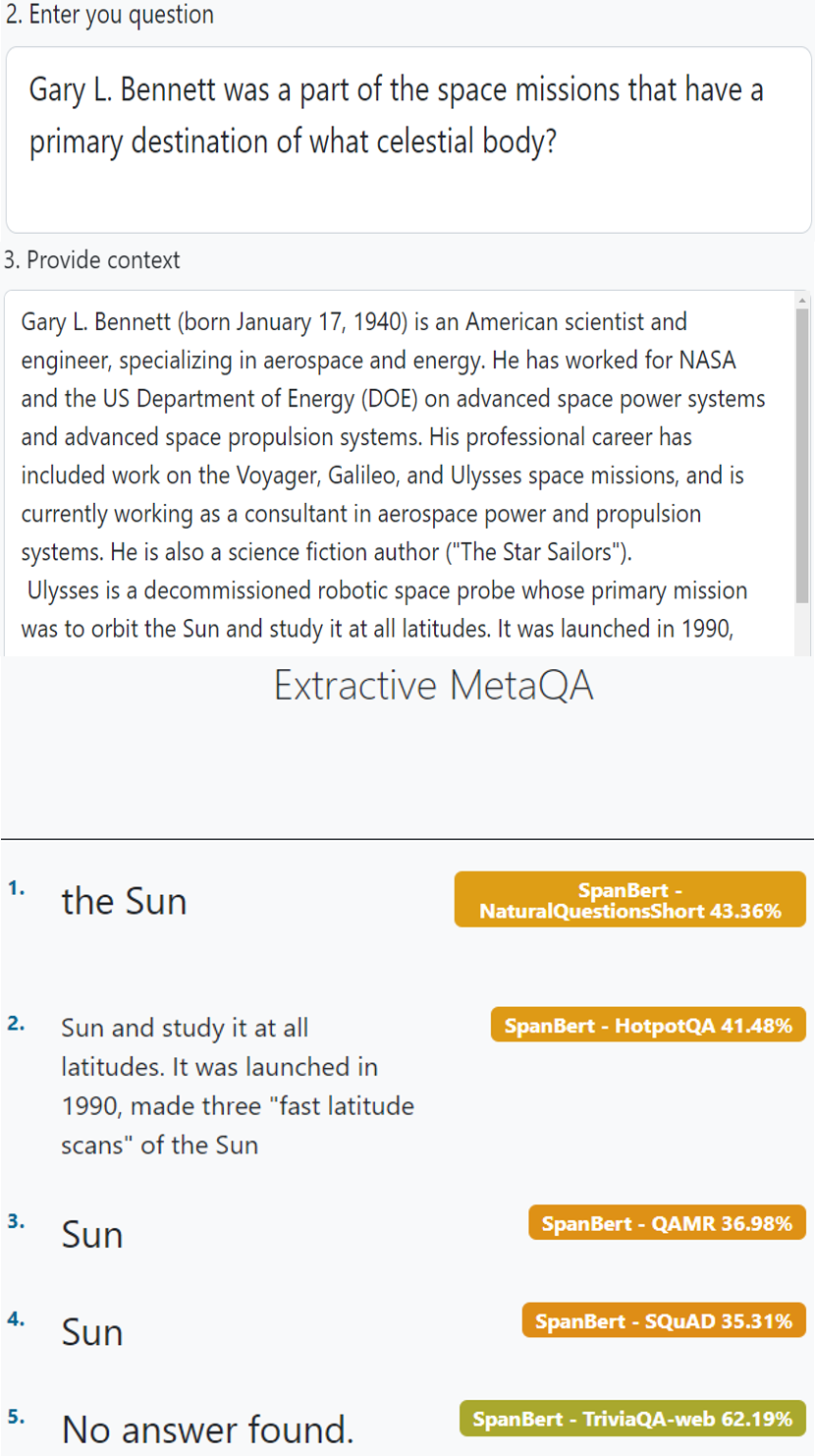}
    \caption{Multi-Agent QA in UKP-SQuARE: different agents are selected to predict the answer based on the input}
    \label{fig:metaqa}
\end{figure}

Thanks to UKP-SQuARE, we can also follow a flipped classroom methodology to teach multi-agent systems. After reading class materials explaining the models of this topic at home, the class time can be used as an explanation of the topic with a live demonstration of these models. In particular, we can easily show that multi-agent systems such as MetaQA \citep{puerto2021metaqa} select different agents depending on the input question. Figure~\ref{fig:metaqa} shows that the first answer selected by MetaQA, which is the correct one, is from an out-of-domain agent, while the second answer, which is not correct, is from the in-domain agent. This example illustrates the collaboration between agents achieved by multi-agent systems and can be an ideal way of starting the lecture on this topic before explaining the architectural details of MetaQA.
Similarly, the platform can be used to introduce multi-dataset systems such as UnifiedQA \citep{khashabi-etal-2020-unifiedqa}, before delving into in-detail explanations of the model. In particular, the lecturer can explain the multiple accepted QA formats by UnifiedQA through real examples, and then, continue the explanation with the training details of the model with the support of slides.

\subsection{Assignments with UKP-SQuARE}

In addition to the above teaching scenarios in class, we also propose a homework assignment based on UKP-SQuARE\footnote{\url{https://colab.research.google.com/drive/17qw1dLWmU5EDxf9TLR29zIG9-EGKmNxP?usp=share_link}} that leverages the insights and knowledge they acquire from the class. The students need to train their own QA model using the popular Hugging Face's Transformer library~\citep{wolf-etal-2020-transformers}, deploy the model on our platform, and then write an in-detail report where they analyze their model from multiple perspectives. This report must include a quantitative analysis of the performance of their model on the test set and also a qualitative analysis that includes an explanation of the outputs of the model to a series of input questions, adversarial attacks that shows errors of their model, and an analysis of the possible behavioral errors obtain from \textit{CheckList}. Furthermore, the students should also compare their model with other available models and identify the type of questions where their model fails. This would help them understand that models overfit the domain of their training data and, therefore, may fail in other domains. This assignment requires students to truly understand each component they learned during the class, which will help them consolidate their knowledge and develop a deeper understanding of the inner workings of different QA techniques. Additionally, the assignment can serve as a useful assessment tool, enabling teachers to gauge students' understanding of the material and provide targeted feedback and support as needed.

\subsection{User Study}
To quantitatively evaluate the effectiveness of UKP-SQuARE in teaching the above QA techniques, we designed a questionnaire to collect feedback from students. The questionnaire was administered to a group of students who had completed a graduate NLP course that used our platform in both class time and for the assignment. All participants are 20-to-30 years-old graduate students in computer science. The questionnaire mainly focuses on two aspects: whether UKP-SQuARE deepens their understanding of techniques in QA systems and whether it makes it easier to get hands-on experience in UKP-SQuARE. The majority of questions require students to rate on a scale of 1 to 5. The complete questionnaire can be found in Appendix ~\ref{app:questionnaire}.

\begin{figure}[!htb]
    \centering
    \includegraphics[width=\columnwidth]{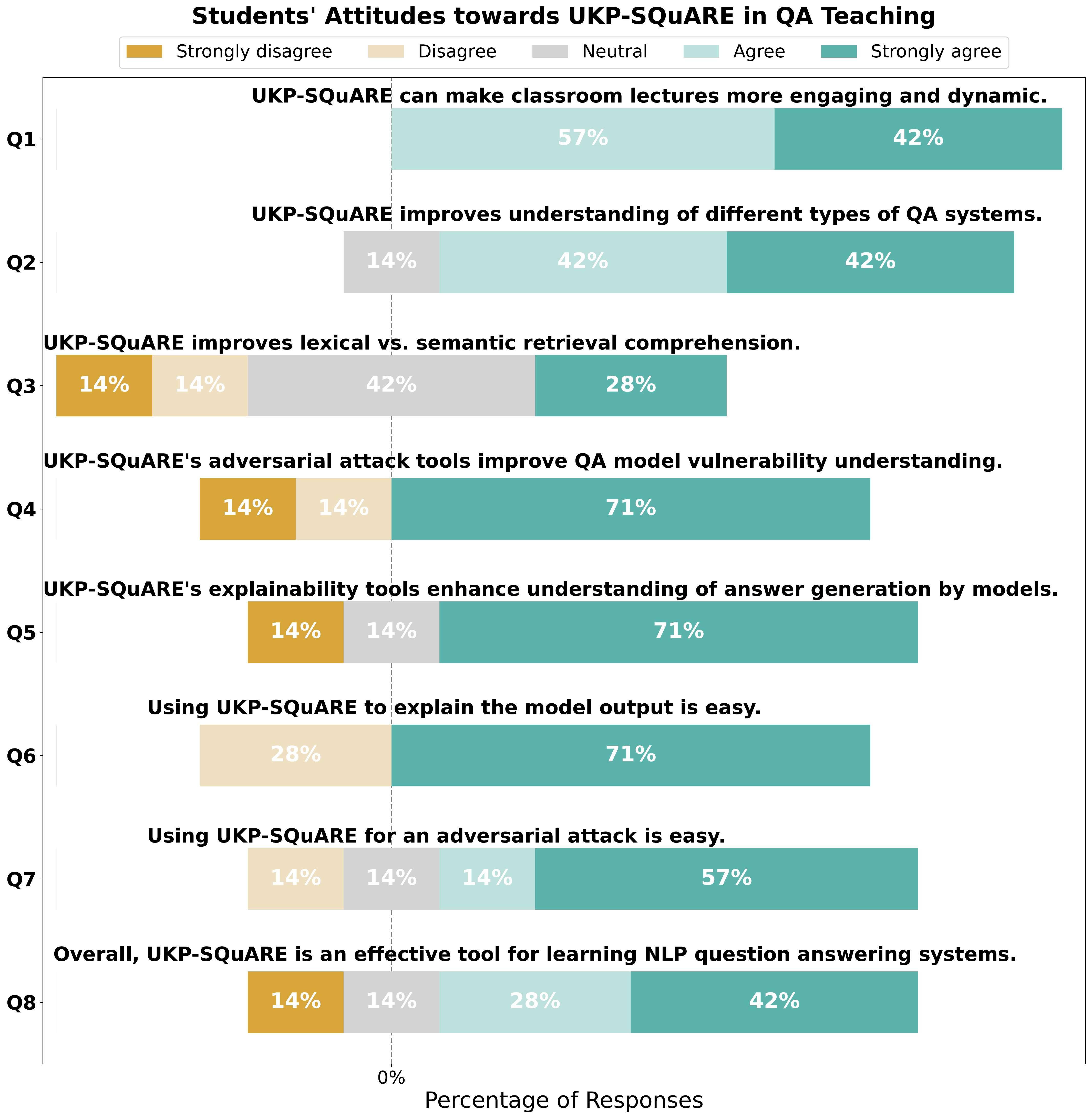}
    \caption{Students feedback towards UKP-SQuARE used in QA education.}
    \label{fig:likert_plot}
\end{figure}

Figure~\ref{fig:likert_plot} shows the Likert scale chart with the responses of seven students who participated in the survey.
As we can see, students have very positive attitudes towards all aspects of UKP-SQuARE for their QA learning.
All participants think that the platform makes the class more engaging and interesting. In particular, most of them (91\%) think UKP-SQuARE helps them better distinguish different QA formats.\
For information retrieval, the majority of the responders do not think that the platform can help them understand better the difference between lexical retrieval and semantic retrieval.
The main reason behind this is that the difference between lexical and semantic retrievers is challenging to distinguish only via visualization unless students actively compare the documents by themselves. Besides, it also requires students to have a good understanding of semantic similarity and lexical similarity. Therefore, we plan to improve it by showing the difference between vector similarity and keyword matching between questions and retrieved documents.
Regarding explainability and adversarial attack tools, around two-thirds of students believe that the platform facilitates their learning process of these topics.
When it comes to hands-on experience, the vast majority of students agree that UKP-SQuARE is easy to use. Our platform provides an infrastructure that dramatically lowers the bar for students to get hands-on experience. All students think that without UKP-SQuARE, they would spend more time finding suitable open-source software to compare different models, analyze the output, and conduct adversarial attacks.
Moreover, the respondents estimated that without UKP-SQuARE, the average time spent on homework would increase from 2-5 hours to more than 8 hours.
One student also commented that doing experiments with the platform was straightforward and allowed him to try different ideas without any overhead. Therefore, although the survey sample is small and limits the conclusions, this overall positive feedback invites us to continue investigating how to conduct our QA and NLP classes more interactively with UKP-SQuARE and suggests that our students would benefit from extending this interactive class to other NLP topics such as generative pre-trained large language models, prompting with reinforcement learning from human feedback, word embeddings, parsing trees, and machine translation among others.

\section{Related Work}
The most relevant tool is the AllenNLP demo\footnote{\url{https://demo.allennlp.org/reading-comprehension/}}, which provides a user interface to the main components of the AllenNLP library~\citep{gardner-etal-2018-allennlp}. This website includes an interface where users can interact with five extractive QA models. However, their goal is to have a showcase of their library rather than an extensive platform for teaching QA. Thus, their functionalities are limited. Most of their deployed models are outdated, only cover extractive QA settings, and do not provide information retrieval methods. Moreover, their explainability and adversarial attacks are not compatible with their transformer-based model. Furthermore, they do not provide graph-based models, which can be useful to explain graph neural networks and explainability methods based on graphs. Additionally, it cannot be used for our homework assignment because users cannot deploy and analyze their own models with explainability and adversarial attack tools as in our platform. However, they do provide demos for other NLP topics, such as Open Information Extraction and named entity recognition, and parsing trees, among others.

\section{Conclusion}
In this paper, we present a novel method to teach question-answering to postgraduate NLP students following the learner-centered method of flipped classrooms. We propose to provide reading materials to the students before the class and use the UKP-SQuARE platform as a driving tool to conduct the class. This platform integrates the most popular QA pipelines and an ecosystem of tools to analyze the available models. These tools include explainability methods, behavioral tests, adversarial attacks, and graph visualizations. We provide a series of use cases for teaching based on the provided models and methods by UKP-SQuARE, showing that classes can become much more interactive by using UKP-SQuARE than in conventional lectures. To evaluate the effectiveness of the platform and our methodology, we conducted a survey to collect feedback from students who took our class. The results show that most of the students think UKP-SQuARE accelerates their learning process and reduces the overhead to get hands-on experience. We plan to extend our platform to support prompting large language models, and therefore, we leave as future work creating a curriculum to teach prompting methods.

\section*{Acknowledgements}
We thank Max Eichler, Martin Tutek, Thomas Arnold, Tim Baumgärtner, and the anonymous reviewers for their insightful comments on a previous draft of this paper. This work has been funded by the German Research Foundation (DFG) as part of the UKP-SQuARE project (grant GU 798/29-1), the QASciInf project (GU 798/18-3), and by the German Federal Ministry of Education and Research and the Hessian Ministry of Higher Education, Research, Science and the Arts (HMWK) within their joint support of the National Research Center for Applied Cybersecurity ATHENE.

\bibliography{anthology,custom}
\appendix

\section{Questionnaire} \label{app:questionnaire}
 The questionnaire includes two parts:
\begin{itemize}
    \item Whether UKP-SQuARE deepens their understanding of QA topic. Some exemplary questions are:
        \begin{itemize}
            \item Does UKP-SQuARE help you understand different types of QA systems better (e.g. extractive QA, abstractive QA)?
            \item Does the adversarial attack tool in UKP-SQuARE help you understand the potential vulnerability of QA models better?
            \item Does the explainability tool in UKP-SQuARE help you understand better how the model generates answers based on the input?
            \item Does using UKP-SQuARE in the classroom make the lecture more dynamic and engaging?
        \end{itemize}
    \item Whether UKP-SQuARE makes it easier to get hands-on experience. Some exemplary questions are:
        \begin{itemize}
            \item How long did you spend on the assignment?
            \item If you don't use UKP-SQuARE, what will you use to finish your assignment (which involves comparing different models, and adversarial attacks)?
            \item Without UKP-SQuARE, how long do you think you need to finish your assignment(including searching for platforms or building a small service by yourself)?
            \item How easy it is to use UKP-SQuARE to do adversarial attacks against models?
            \item How easy it is to use UKP-SQuARE to explain the model output?
            \item If you don't use UKP-SQuARE and you need to perform adversarial attacks on your model, would you be able to complete the assignment? If so, how much more difficult would it be? 
            \item If you don't use UKP-SQuARE and you need to interpret the answers of your model using saliency maps, would you be able to do it? if so, how much more difficult would it be?
            \item Does UKP-SQuARE UI help you compare models easier? (eg: compared to using Jupyter Notebooks)?
        \end{itemize}    
            
\end{itemize}

\end{document}